\documentclass{article} 
\usepackage[final]{COLM2025/colm2025_conference}

\usepackage{microtype}
\usepackage{hyperref}
\usepackage{url}
\usepackage{booktabs}

\usepackage{lineno}

\usepackage{amsthm, amsmath, amssymb, bbm, mathabx}
\allowdisplaybreaks 
\usepackage{graphicx}
\usepackage{enumerate}
\usepackage{pifont}
\usepackage{booktabs}
\usepackage{multirow}
\usepackage{multicol}
\usepackage{stfloats}
\usepackage{xspace}
\usepackage{thmtools}
\usepackage{thm-restate}
\usepackage{hyperref}
\usepackage{cleveref}
\usepackage{anyfontsize} 
\usepackage{makecell}
\usepackage{hhline}
\usepackage{colortbl}
\usepackage{xpatch}
\usepackage{scalerel,stackengine}
\usepackage{sidecap}
\usepackage{float}
\usepackage{mdframed}
\usepackage{listings}
\usepackage{pdfpages}
\usepackage{lscape}

\makeatletter
\def\@fnsymbol#1{\ensuremath{\ifcase#1\or *\or \dagger\or \ddagger\or
  \mathsection\or \mathparagraph\or \|\or \diamond \or **\or \dagger\dagger
  \or \ddagger\ddagger \else\@ctrerr\fi}}
\makeatother

\makeatletter
\newcommand{\printfnsymbol}[1]{%
  \textsuperscript{\@fnsymbol{#1}}%
}
\makeatother

\newcommand{\para}[1]{\paragraph{#1}}

\crefname{condition}{Condition}{Conditions}

\crefname{assumption}{Assumption}{Assumptions}

\theoremstyle{definition}

\newfloat{textbox}{tbp}{lob}
\floatname{textbox}{Text Box}
\crefname{textbox}{Text Box}{Text Boxes}

\newmdenv[
  linewidth=1pt,
  skipabove=10pt,
  skipbelow=10pt,
  backgroundcolor=gray!10,
  roundcorner=5pt,
  leftmargin=10pt,
  rightmargin=10pt
]{textframe}

\lstdefinelanguage{text}{
  basicstyle=\ttfamily\scriptsize\selectfont\color[rgb]{0,0,0},
  keywordstyle=\color[rgb]{1, 0., 0.},
  sensitive=true,
  morecomment=[l]{//},
  morecomment=[s]{/*}{*/},
  commentstyle=\color[gray]{0.4},
  morestring=[b]",
  morestring=[b]',
  columns=fullflexible,
  numbers=none,
  framerule=0pt,           
  framexleftmargin=0em,    
  framexrightmargin=0em,   
  xleftmargin=0pt,
  xrightmargin=0pt,        
  frame=none,
  breaklines=true,
  framesep=0pt,            
  resetmargins=true        
}

\newcommand{\floor}[1]{\left\lfloor #1 \right\rfloor}

\newcommand{\abs}[1]{\left| #1 \right|}

\newcommand{\ov}[1]{\overline{#1}}
\newcommand{\ud}[1]{\underline{#1}}

\newcommand{\cD}{\mathcal{D}}

\newcommand{\cF}{\mathcal{F}}

\newcommand{\cK}{\mathcal{K}}
\newcommand{\cL}{\mathcal{L}}
\newcommand{\cM}{\mathcal{M}}

\newcommand{\cQ}{\mathcal{Q}}

\newcommand{\cX}{\mathcal{X}}

\newcommand{\bbE}{\mathbb{E}}

\newcommand{\ee}{\textup{e}}

\newcommand{\softmax}{\mathsf{Softmax}}

\newcommand{\Roma}[1]{\uppercase\expandafter{\romannumeral#1}}

\newcommand{\Fwiki}{\ensuremath{\cF_{\mathsf{wiki}}}\xspace}
\newcommand{\Fts}{\ensuremath{\cF_{\mathsf{ts}}}\xspace}
\newcommand{\Fcode}{\ensuremath{\cF_{\mathsf{code}}}\xspace}
\newcommand{\fwiki}{\ensuremath{f_{\mathsf{wiki}}}\xspace}
\newcommand{\fts}{\ensuremath{f_{\mathsf{ts}}}\xspace}
\newcommand{\fcode}{\ensuremath{f_{\mathsf{code}}}\xspace}

\newcommand{\Kwiki}{\ensuremath{\cK_{\mathsf{wiki}}}\xspace}
\newcommand{\Kts}{\ensuremath{\cK_{\mathsf{ts}}}\xspace}

\newcommand{\kwiki}{\ensuremath{k_{\mathsf{wiki}}}\xspace}
\newcommand{\kts}{\ensuremath{k_{\mathsf{ts}}}\xspace}
\newcommand{\kcode}{\ensuremath{k_{\mathsf{code}}}\xspace}

\newcommand{\Qwiki}{\ensuremath{\cQ_{\mathsf{wiki}}}\xspace}
\newcommand{\Qts}{\ensuremath{\cQ_{\mathsf{ts}}}\xspace}

\newcommand{\qwiki}{\ensuremath{q_{\mathsf{wiki}}}\xspace}
\newcommand{\qts}{\ensuremath{q_{\mathsf{ts}}}\xspace}
\newcommand{\qcode}{\ensuremath{q_{\mathsf{code}}}\xspace}

\newcommand{\Dwiki}{\ensuremath{\cD_{\mathsf{wiki}}}\xspace}
\newcommand{\Dts}{\ensuremath{\cD_{\mathsf{ts}}}\xspace}
\newcommand{\Dcode}{\ensuremath{\cD_{\mathsf{code}}}\xspace}

\newcommand{\Nocc}{\ensuremath{N_{\mathsf{occ}}}\xspace}
\newcommand{\Noccx}{\ensuremath{N_{\mathsf{occ}}^{\mathsf{x}}}\xspace}
\newcommand{\Nocctest}{\ensuremath{N_{\mathsf{occ}}^{\mathsf{test}}}\xspace}

\newcommand{\Lc}{\ensuremath{L_{\mathsf{ctx}}}\xspace}
\newcommand{\Lb}{\ensuremath{L_{\mathsf{blk}}}\xspace}
\newcommand{\Lk}{\ensuremath{L_{\mathsf{knw}}}\xspace}

\newcommand{\knw}{\emph{knowledge}\xspace}

\newcommand{\red}[1]{\textcolor{red!80}{#1}}

\usepackage[colorinlistoftodos, textwidth=18mm]{todonotes}

\def\shownotes{1}
\ifnum\shownotes=0
\newcommand{\todorz}[1]{}
\newcommand{\todorzout}[1]{}
\newcommand{\todossdout}[1]{}
\newcommand{\todossd}[1]{}
\newcommand{\todoruizhe}[1]{}
\else
\newcommand{\todorz}[1]{\todo[color=blue!10, inline]{\small RZ: #1}}
\newcommand{\todorzout}[1]{\todo[color=blue!10]{\scriptsize RZ: #1}}
\newcommand{\todossdout}[1]{\todo[color=red!10]{\scriptsize SSD: #1}}
\newcommand{\todossd}[1]{\todo[color=red!10, inline]{\small SSD: #1}}
\newcommand{\todoruizhe}[1]{\todo[color=purple!10, inline]{\small Ruizhe: #1}}
\fi

\definecolor{darkblue}{rgb}{0, 0, 0.5}
\hypersetup{colorlinks=true, citecolor=darkblue, linkcolor=darkblue, urlcolor=darkblue}

\title{CASCADE Your Datasets for\\Cross-Mode Knowledge Retrieval of Language Models}

\author{
Runlong Zhou\thanks{Part of this work done when Runlong was an intern at Microsoft Research, Redmond.}\\
University of Washington \\
\texttt{vectorzh@cs.washington.edu} \\
\And
Yi Zhang\\
xAI \\
\texttt{zhayi0928@gmail.com}
}

\begin{document}

\ifcolmsubmission
\linenumbers
\fi

\maketitle

\begin{abstract}
Language models often struggle with cross-mode knowledge retrieval -- the ability to access knowledge learned in one format (mode) when queried in another.
We demonstrate that models trained on multiple data sources (e.g., Wikipedia and TinyStories) exhibit significantly reduced accuracy when retrieving knowledge in a format different from its original training mode.
This paper quantitatively investigates this phenomenon through a controlled study of random token sequence memorization across different modes.
We first explore dataset rewriting as a solution, revealing that effective cross-mode retrieval requires prohibitively extensive rewriting efforts that follow a sigmoid-like relationship.
As an alternative, we propose CASCADE, a novel pretraining algorithm that uses cascading datasets with varying sequence lengths and computing losses on only the second half of each training sequence to capture knowledge at different scales.
Our experiments demonstrate that CASCADE outperforms dataset rewriting approaches, even when compressed into a single model with a unified loss function.
This work provides both qualitative evidence of cross-mode retrieval limitations and a practical solution to enhance language models' ability to access knowledge independently of its presentational format.
To facilitate research in the field of LLMs, the code is publicly released.\footnote{\url{https://github.com/zhourunlong/CASCADE\_public}}
\end{abstract}

\section{Introduction}

\begin{figure}[htbp]
    \centering
    \vspace{-0.4cm}
    \includegraphics[width=0.98\linewidth]{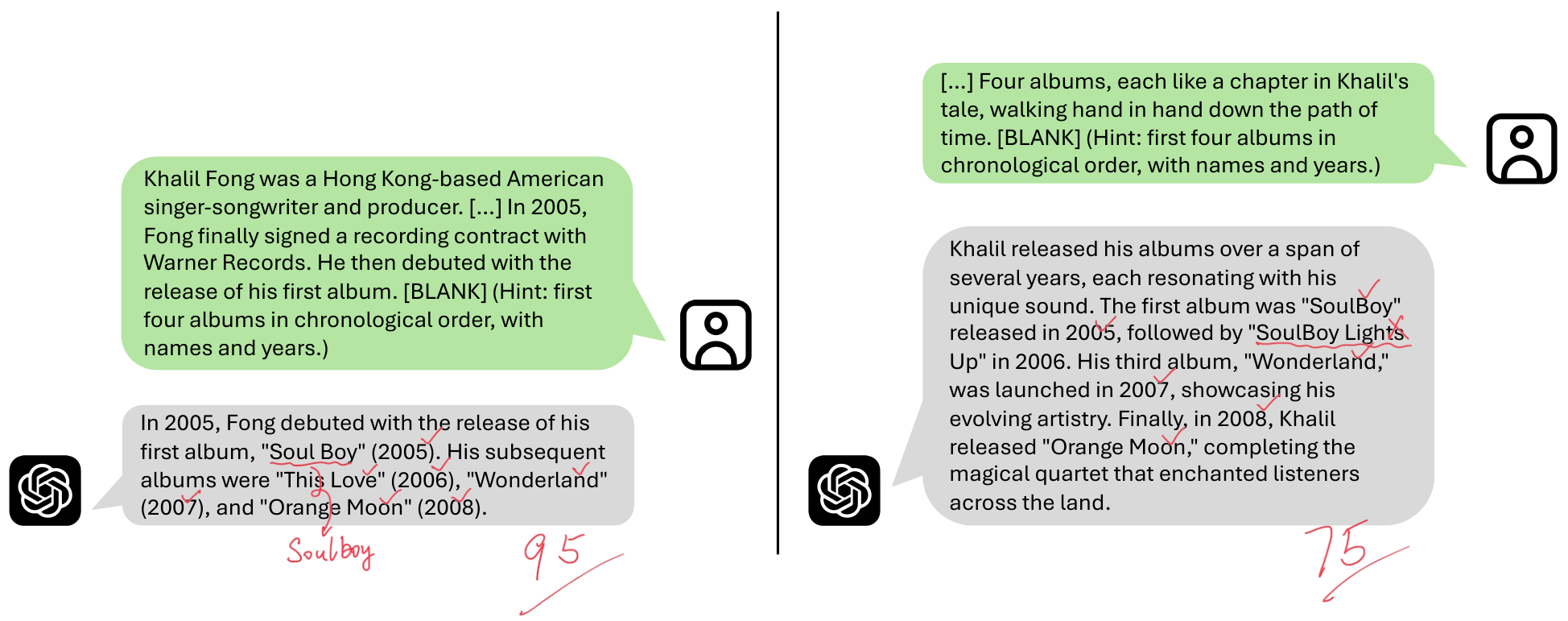}
    \vspace{-1em}
    \caption{GPT-4o shows inconsistent accuracies when prompted with the same question but in different formats. 
    Left: the query is in a Wikipedia format. Right: the query is in a story format.
    Find more detailed examples in \Cref{tab:gen1,tab:gen2,tab:gen3}. }
    \label{fig:qualitative}
\end{figure}

Large language models (LLMs) are often pretrained on corpus comprised of several sources, each with a unique \textbf{\emph{mode}} (wording style, organization format, etc., will also be referred to as \textbf{\emph{format}}).
Although LLMs can achieve low losses on validation sets from the same mode, we observe concrete examples that they cannot perform \textbf{\emph{cross-mode knowledge retrieval}} effectively.
For example, we can pretrain a language model on both Wikipedia excerpts and the TinyStories \citep{eldan2023tinystoriessmalllanguagemodels} dataset until convergence.
However, when we query the model for knowledge present in the Wikipedia training set \emph{using a story format}, the generated response shows surprisingly low accuracy on average.
We illustrate this in \Cref{fig:qualitative}, with details in \Cref{sec:qualitative_gen}.

Motivated by this phenomenon, we research the following question:
\begin{center}
\textit{\textbf{How can we make language models capable of cross-mode knowledge retrieval?}}
\end{center}
We approach this problem quantitatively, focusing on a toy yet fundamental task of memorizing \textbf{\emph{random token sequences}} in different modes (Wikipedia and TinyStories).
Memorization of random token sequences can be precisely quantified by computing log probabilities.
We investigate whether language models learn \emph{spurious correlations} between \textbf{\emph{knowledge}} and \textbf{\emph{mode}} instead of learning knowledge independently.
To the best of our knowledge, while spurious correlations in natural language processing have been widely studied in classification tasks, \emph{\textbf{they remain underexplored in general language modeling tasks, particularly in knowledge memorization and manipulation}}.
We hope this work will serve as an initial study of spurious correlations in general language modeling and inspire more effective and efficient methods to alleviate this issue.

\subsection{Our contributions}

Our contributions are twofold - both qualitative and quantitative.

\textbf{Qualitatively}, we build a pipeline (\Cref{sec:qualitative_gen}) that demonstrates how LLMs fail at \textbf{\emph{cross-mode knowledge retrieval}}.

\textbf{Quantitatively}, we focus on the \emph{pretraining} stage to improve language models' \textbf{\emph{cross-mode knowledge retrieval}} capability.
Our quantitative contributions can be summarized as follows:

$\bullet$ \textbf{Dataset rewriting.} We first study how the \emph{ratio} between non-cross-mode (original in the dataset) and cross-mode data (\emph{rewritten} into the dataset, with occurrences controlled by us) affects the evaluation performance of cross-mode knowledge retrieval (\Cref{sec:rewrite}).
We plot curves of evaluation performance with respect to the ratio $r$ between same-mode and cross-mode knowledge occurrences.
These curves follow a \emph{sigmoid}-like function: $f (r) = a \cdot \sigma (b (\log (r) - c))$.
These results demonstrate that \textbf{\emph{effective cross-mode knowledge retrieval requires extensive rewriting effort, which is prohibitive in practice}}.

$\bullet$ \textbf{Novel algorithm: CASCADE.} We propose a novel algorithm, CASCADE, as a solution.
During pretraining, we use a series of \textbf{\emph{cascading}} datasets with \emph{different sequence lengths} to help the language model \emph{capture knowledge at different scales}.
The loss is calculated on only the second half of each training sequence.
We first show that an original form of CASCADE using model ensemble achieves better performance than dataset rewriting (\Cref{sec:original_cascade}), then improve its complexity by compressing it into a single model with a single loss function (\Cref{sec:cascade}).
We also visualize how different sequence lengths contribute to completing different knowledge.


\section{Related works}

We discuss the most related line of works here, deferring the other works to \Cref{sec:additional_related_works}.

\para{Consistency.}
Consistency in language models has earned significant research attention.
\citet{elazar2021measuring} defined consistency as "invariance under meaning-preserving alternations" and introduced PARAREL for evaluating factual knowledge consistency across paraphrases.
Inconsistency manifests across various NLP applications: \citet{ribeiro-etal-2019-red} identified inconsistencies in question answering systems, while \citet{kryscinski2019evaluating} studied factual consistency in summarization.
\citet{li2019logic} and \citet{camburu2019make} examined inconsistencies in natural language inference (NLI) systems and explanations, respectively.
Researchers have proposed various improvement approaches: \citet{elazar2021measuring} introduced a consistency loss function, \citet{kassner2021enriching} proposed augmenting PLMs with an evolving memory, \citet{chen2021can} developed explanation-based post-training, and \citet{asai2020logic} utilized data augmentation with symmetricity properties.

\emph{We highlight that while at a high level the issues associated with \textbf{cross-mode knowledge retrieval} could be classified as inconsistency, they differ drastically.
In previous consistency studies, input changes are typically small perturbations such as synonym replacement, word or sentence permutation, or statement-QA conversion, leaving word styles largely unchanged.
In contrast, \textbf{cross-mode knowledge retrieval} applies to entirely different text sources with highly diverse word styles, making the language model more prone to derive spurious correlations between the mode and the knowledge.}
\section{Settings} \label{sec:settings}

In this work, we study the knowledge memorization mechanism in language models.
Specifically, we care about \textbf{\emph{how much will the format text influence the language model's memorization of the knowledge}}, and \textbf{\emph{how to reduce this influence}}.
To this end, we will construct datasets that admit a well-defined criterion of the extent of memorization.
The high-level idea is to define knowledge pieces as \emph{random token sequences}, thus any language model is said to memorize the knowledge only if it can perfectly \emph{generate the whole sequences}, admitting log probabilities as quantification of memorization.
The \emph{modes} or \emph{formats} are defined as texts from different datasets.
The language models should separate knowledge from modes to perform well on cross-mode knowledge retrieval tasks.

\para{Tokenization.}
We process everything in the \emph{token space}.
We use the GPT-2 tokenizer (\texttt{tiktoken.get\_encoding("gpt2")}) in this study, which has a token range of $[0, 50256]$.
Some other tokens may be used in the experiments, and we constrain them to be in a \emph{separate} range of $[50257, 50303]$.

\para{Notations.}
Denote $\Sigma$ as the set of all possible tokens.
We use subscripts to denote the mode name, superscripts to denote the index in a set, and numbers in brackets to denote the index in a set.
For a set $\cX$, we use $\abs{\cX}$ to denote the number of unique elements in $\cX$.
For a sequence $a$, we use $\abs{a}$ to denote the length of $a$.

\para{Indexing.}
We follow Python's indexing convention.
Numerical indices start from $0$.
When using a range to index, the lower bound is included while the upper bound is \emph{excluded}.
When indexing an array $a$, a lower bound of $0$ or an upper bound of $\abs{a}$ can be omitted.
A \emph{negative} index $a[-i]$ means $a[\abs{a} - i]$.

\subsection{Core concepts}

First, we introduce core concepts that will be referenced frequently when constructing datasets and during training and evaluation.

\para{Formats/Modes.}
We use existing datasets, English Wikipedia excerpts and TinyStories \citep{eldan2023tinystoriessmalllanguagemodels}, as \emph{format/mode texts}.
They are denoted as \Fwiki and \Fts, respectively.
For training, we take portions from each format, making them roughly equal in token counts.
We take disjoint portions from each format for evaluation.

\para{Knowledge.}
We use \textbf{\emph{random token sequences}} as \emph{knowledge} for the following reasons:

$\bullet$ \textbf{Quantification:} Memorizing random token sequences requires precise token-by-token memorization, unlike general knowledge which can be rephrased in various ways.
This enables exact quantification by computing the log probability of generating a desired random token sequence.

$\bullet$ \textbf{Exclusiveness:} We can ensure these knowledge pieces neither appear in mode texts nor correlate with each other.
This prevents knowledge leakage in the training set and eliminates correlation between mode and knowledge.

We construct $K=32$ pieces of knowledge for each mode:
\begin{align*}
    \Kwiki = \{\kwiki^{(0)}, \kwiki^{(1)}, \ldots, \kwiki^{(K-1)}\}, \quad \textup{and} \quad \Kts = \{\kts^{(0)}, \kts^{(1)}, \ldots, \kts^{(K-1)}\}.
\end{align*}
Each piece of knowledge $k \in \Kwiki \cup \Kts$ is a random token sequence with length between $\ud{\Lk}=8$ and $\ov{\Lk}=512$ (both inclusive), and the tokens are from the range of $[50296, 50303]$.
Each position in the sequence is independently sampled from a uniform distribution over the token range.
To make knowledge exclusive to its corresponding format, these two knowledge sets are \emph{disjoint at the sequence level}: $\Kwiki \cap \Kts = \varnothing$.

\para{Queries.}
Queries are ``hints'' for the language model to complete a knowledge piece, so we set them as \emph{prefixes} of each knowledge piece.
To make the problem well-defined, the prefixes should be unique so that they correspond to knowledge pieces in a one-to-one manner.
We find the shortest prefix length so that the induced queries are different:
\begin{align*}
    \ell = \min l \quad \textup{such that} \quad \abs{\{k[0:l] \mid k \in \Kwiki \cup \Kts\}} = 2 K.
\end{align*}
The queries are defined as
\begin{align*}
    \Qwiki = \{\qwiki^{(i)} := \kwiki^{(i)} [0:\ell] \mid 0\le i < K\}, \quad \textup{and} \quad \Qts = \{\qts^{(i)} := \kts^{(i)} [0:\ell] \mid 0\le i < K\}.
\end{align*}

\subsection{Problem formulation} \label{sec:prob_formulation}

Now we formally describe our problem of interest: \textit{\textbf{cross-mode knowledge retrieval}}.

\begin{figure}[t]
    \centering
    \vspace{-0.2cm}
    \includegraphics[scale=0.5]{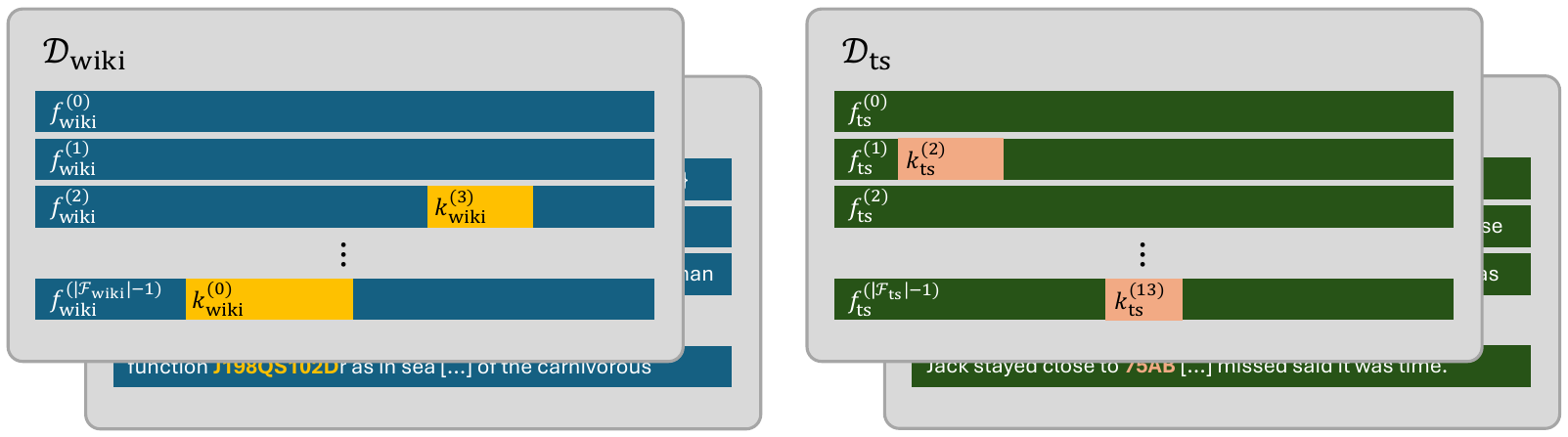}
    \vspace{-1em}
    \caption{Illustration of the datasets. Each line represents a block of \Lb consecutive tokens in \Fwiki or \Fts. Knowledge pieces overwrite to some of the blocks in arbitrary positions. The de-tokenized datasets are shown in the background.}
    \label{fig:dataset}
\end{figure}

\para{Datasets.}
There are two fixed datasets, \Dwiki and \Dts, each containing knowledge from only one mode (itself).
Taking Wikipedia as an example:
The format texts from \Fwiki are divided into consecutive blocks with length $\Lb=1024$.
We set hyperparameter $\Nocc = 8192$ as the number of occurrences\footnote{This satisfies the $1000$-exposure requirement in \cite{allenzhu2024physicslanguagemodels33}.} for each knowledge piece $k \in \Kwiki$ in \Fwiki, totaling $K \Nocc$ occurrences of knowledge pieces.
These knowledge pieces are distributed across $K \Nocc$ blocks sampled uniformly.
Inside each block $\fwiki$, the knowledge piece overwrites a random, consecutive subsequence with equal probability.
An illustration is shown in \Cref{fig:dataset}.

\begin{figure}[htbp]
    \centering
    \includegraphics[scale=0.5]{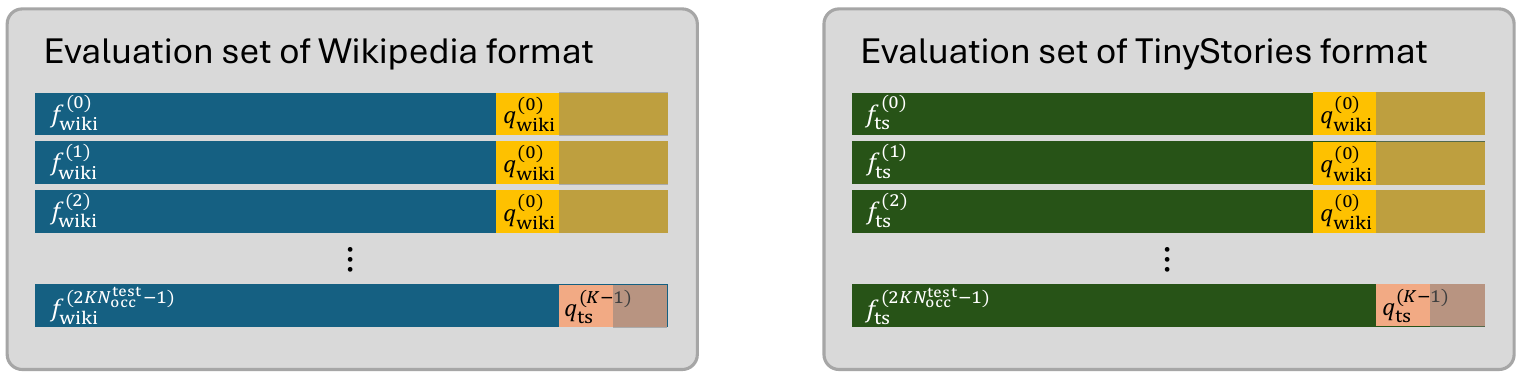}
    \caption{Illustration of the evaluation datasets. The shadowed parts are for completion and log probability calculation. The format texts are taken from the \emph{evaluation} split of \Fwiki and \Fts, so they are different from those in \Cref{fig:dataset}.}
    \label{fig:eval_set}
\end{figure}

\para{Evaluation.}
Given these two datasets, we want to quantify the \textit{\textbf{cross-mode knowledge retrieval}} capability of language models.
Since the only way to memorize the random sequences is to perfectly generate them, the task is to do \emph{completion} on the remaining tokens given a query (``hint'') $q$.
We set $\Nocctest=16$ occurrences for each query $q \in \Qwiki \cup \Qts$.
For each $q$, we randomly sample $\Nocctest$ blocks of length \Lb from the evaluation portion of \emph{each} format, $\Fwiki$ and $\Fts$, and overwrite it to the \emph{end} of each block.
Suppose $q = k[0:\ell]$ for some $k \in \Kwiki \cup \Kts$ by our construction, $\abs{k} = \Lk$, and the format text block is $f$ such that $\abs{f} = \Lb$.
The criteria is the normalized log probability of the completion part:
\begin{align*}
    \frac{1}{\Lk - \ell} \sum_{i=\ell}^{\Lk - 1} \log \cM_\theta (k [i] \mid f[:-\Lk], k[:i]),
\end{align*}
where $\cM_\theta$ is the model parameterized by $\theta$.
An illustration can be found in \Cref{fig:eval_set}.

\section{A straightforward approach: rewrite the datasets} \label{sec:rewrite}

Direct training on $\Dwiki \cup \Dts$ yields poor performance -- as shown in \Cref{fig:ratio}, where dashed horizontal lines represent normalized log probabilities of completions after direct training using $\Dwiki \cup \Dts$.
Qualitative results (\Cref{sec:qualitative_gen}) also support this argument.
The language model likely learned a spurious correlation between mode and knowledge, so when queried with \fwiki\qts or \fts\qwiki, it fails to correctly complete with \kts or \kwiki.

\subsection{Method description}

\begin{figure}[htbp]
    \centering
    \includegraphics[scale=0.5]{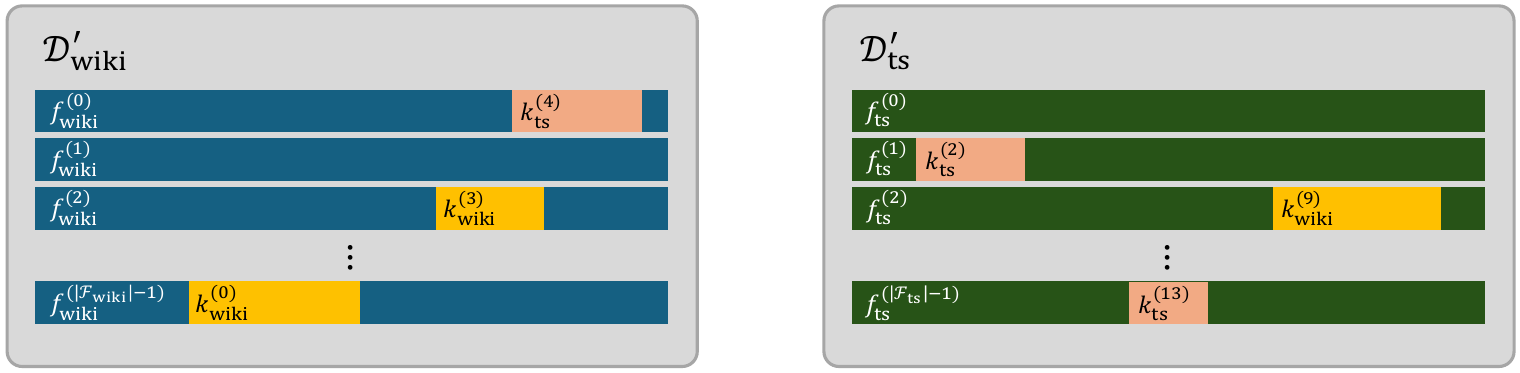}
    \caption{Illustration of dataset rewriting. Readers can compare with \Cref{fig:dataset}.}
    \label{fig:rewrite}
\end{figure}

A straightforward approach to reduce this spurious correlation is to rewrite the datasets, incorporating \emph{cross-mode} knowledge.
For example, when rewriting \Dwiki into $\Dwiki'$, besides the original $K \Nocc$ occurrences of knowledge pieces in \Kwiki, we set a hyperparameter \Noccx as the number of occurrences for each cross-mode knowledge in this dataset.
In practice, identifying and rewriting all exclusive knowledge is costly, so we use only ${\kts^{(0)}, \ldots, \kts^{(K/2-1)}}$ to rewrite the dataset and use ${\kts^{(K/2)}, \ldots, \kts^{(K-1)}}$ as \textbf{\emph{hold-out}} knowledge for evaluation.
Each $\kts \in {\kts^{(0)}, \ldots, \kts^{(K/2-1)}}$ appears exactly \Noccx times in $\Dwiki'$, using the same method to generate \Dwiki.
An illustration can be found in \Cref{fig:rewrite}.

For notational ease, we use the following shorthand:

$\bullet$ \fts\qts and \fwiki\qwiki: evaluation data with a query from the same mode as the format text.

$\bullet$ \fts\qwiki and \fwiki\qts: evaluation data with a cross-mode query.

For example, in \Cref{fig:eval_set}, the first and last entries in the left part are denoted as \fwiki\qwiki and \fwiki\qts, respectively.


\subsection{Results} \label{sec:ratio_results}

\begin{figure}[t]
    \centering
    \vspace{-0.3cm}
    \includegraphics[width=0.98\linewidth]{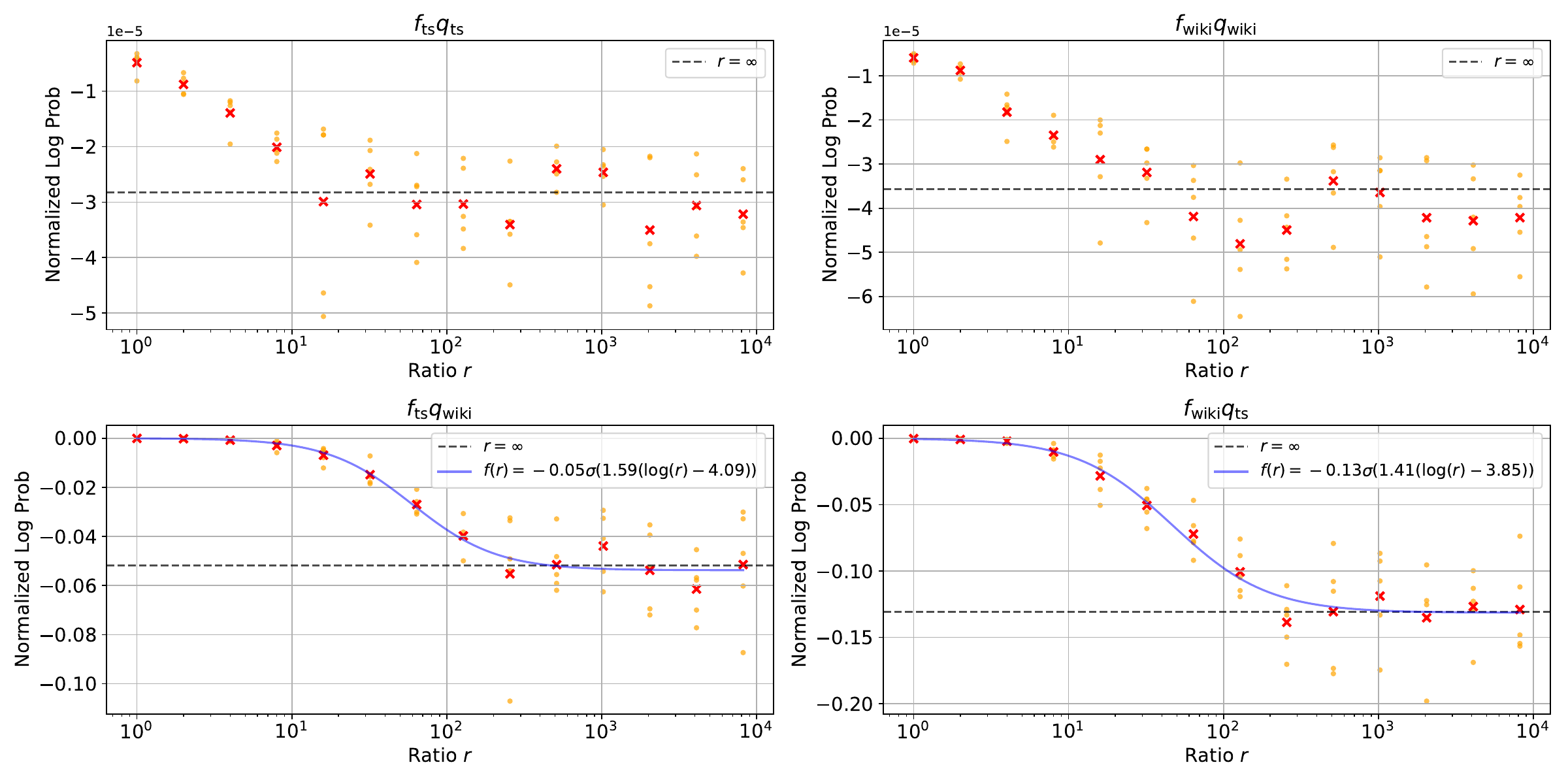}
    \vspace{-1em}
    \caption{Normalized log probabilities for different ratios. The $x$-axis is in log scale. Yellow dots represent results from individual runs with $5$ random seeds, while red crosses show average values. Dashed horizontal lines indicate results from direct training on the original datasets, \Dwiki and \Dts. Cross-mode evaluations use only hold-out queries.}
    \label{fig:ratio}
\end{figure}

We test this method's effectiveness by sweeping over $\Noccx \in \{0\} \cup \{2^i \mid 1 \le i \le 13\}$.
With ratio $r = \Nocc / \Noccx$, we plot the relationship (\Cref{fig:ratio}) between $r$ and the convergent values of normalized log probabilities in evaluation.
Experiment details are deferred to \Cref{sec:quantitative_exp}.
A special case is $r = \infty$ (dashed horizontal lines), corresponding to $\Noccx = 0$, which represents the scenario without rewriting.

We also report in \Cref{tab:ratio} the normalized log probabilities for small ratios $r \in \{ 1.0, 2.0, 4.0 \}$.

\begin{table}[ht]
    \centering
    \begin{tabular}{c!{\vrule width 1.25pt}c|c!{\vrule width 1pt}c|c}
        \noalign{\hrule height 1.5pt}
        & \fts\qts & \fwiki\qwiki & \fts\qwiki & \fwiki\qts \\
        \noalign{\hrule height 1.25pt}
        $r = 1.0$ & $-4.87 \times 10^{-6}$ & $-5.94 \times 10^{-6}$ & $-6.75 \times 10^{-5}$ & $-2.98 \times 10^{-4}$ \\ \hline
        $r = 2.0$ & $-8.78 \times 10^{-6}$ & $-8.80 \times 10^{-6}$ & $-1.93 \times 10^{-4}$ & $-9.25 \times 10^{-4}$ \\ \hline
        $r = 4.0$ & $-1.39 \times 10^{-5}$ & $-1.82 \times 10^{-5}$ & $-7.76 \times 10^{-4}$ & $-2.21 \times 10^{-3}$ \\
        \noalign{\hrule height 1.5pt}
    \end{tabular}
    \caption{Normalized log probabilities for small ratios, averaged over $5$ random seeds.}
    \label{tab:ratio}
\end{table}

\subsection{Remarks} \label{sec:rewrite_remark}

We now make several remarks on the method of dataset rewriting.

$\bullet$ \textbf{The relation between the log ratio and the cross-mode evaluation results roughly follows a sigmoid function.}
We observe that a sigmoid-like function fits the points well, so we perform regression using:
\begin{align*}
    f(r) = a \cdot \sigma (b (\log (r) - c)), \quad \textup{where} \quad \sigma(x) = \frac{1}{1 + \ee^{-x}}.
\end{align*}
The blue curves in \Cref{fig:ratio} display the regressed functions.

$\bullet$ \textbf{Meaningful results only come with extensive rewriting.}
\Cref{tab:ratio} shows that to achieve cross-mode query performance comparable to non-cross-mode queries, the ratio should be at most $4.0$, meaning $\Noccx \ge 2048$.
However, even when $r = 1.0$ ($\Noccx = 8192$), normalized log probabilities for cross-mode queries remain at order $10^{-4}$, still one order of magnitude worse than non-cross-mode queries.
Additionally, we rewrote half of the different knowledge pieces, resulting in rewritten knowledge of the same order as the original knowledge.
In practice, such extensive rewriting requires significant human effort to identify and rewrite knowledge differently across contexts.
Even with the help of language models, rewriting prompts need case-by-case design, which also demands substantial human effort.
Moreover, we \emph{hypothesize} that:
\begin{center}
    \textbf{\emph{Data with $n \gg 2$ independent modes requires rewriting all $n (n-1) / 2$ pairs.}}
\end{center}
If this hypothesis is true, dataset rewriting becomes prohibitively impractical for achieving satisfactory performance.
Results in \Cref{tab:3_modes} support this hypothesis when $n = 3$.

\section{A cure: CASCADE the datasets} \label{sec:full_cascade_content}

As a starting point, we consider an easier problem: suppose the knowledge can only appear in the \emph{end} of each sequence blocks of length $\Lb$, and they all have the \emph{same} lengths of $\Lk$.
Assume that $\Lb$ is a multiple of $\Lk$.
If we want the model to perfectly memorize the knowledge without being affected by modes, we can use a context length of $\Lc = \Lk$ in training.
This guarantees that each piece of knowledge fits exclusively in some training sequence, so that \emph{it is not correlated with any mode}.

In the problem described in \Cref{sec:prob_formulation}, we know neither the exact position nor the exact length of knowledge pieces, making it impossible to fit them exclusively within training sequences.
As an alternative design, we aim to ensure each knowledge piece occupies \emph{a large portion} of some training sequence to minimize the influence of modes.

\subsection{Capturing knowledge with doubling context lengths} \label{sec:original_cascade}

\begin{figure}[t]
    \centering
    \vspace{-0.6cm}
    \includegraphics[width=0.6\textwidth]{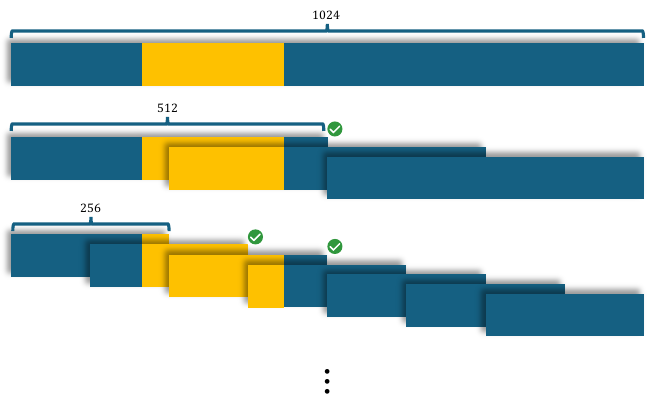}
    \vspace{-1em}
    \caption{Illustration of \emph{\textbf{cascading}} datasets. The highlighted part represents the knowledge. Blocks with a check mark in the top right indicate that the corresponding sequence captures the knowledge, satisfying \Cref{eq:sol_all_req}.}
    \label{fig:cascade}
\end{figure}

Roughly speaking, for a knowledge piece of length $\Lk$, if we set the context length $\Lc \le 2 \Lk$, then regardless of its location in $\cD$, it will occupy \emph{at least half} of the tokens in some training sequence.
This can be guaranteed when training sequences overlap by $\Lc / 2$.
Since we assume $\Lk \le \ov{\Lk} = 512$, we can train a small number of language models with context lengths $8, \ldots, 1024 = \Lc$ using a series of \emph{\textbf{cascading}} datasets (\Cref{fig:cascade}, with details explained in \Cref{sec:cascade_training}).
This ensures each knowledge piece is captured by \emph{at least one} language model.

During evaluation and generation, we predict the next token using a probability distribution that is a weighted average over all models (after normalization).

Since one pass of length $L$ in a transformer requires $\Theta (L^2)$ time, and our context lengths follow a geometric sequence, using all models adds little computational overhead compared to using a single model.
We elaborate on this idea in \Cref{sec:time_complexity}.

\subsubsection{Training} \label{sec:cascade_training}

Now we present a novel algorithm, \emph{\textbf{Original CASCADE}}, realizing the above high-level idea.
``Original'' here is to distinguish it from the compressed, more practical variant that will be introduced in \Cref{sec:cascade}.

We abuse the notation that the original dataset $\cD$ is an array of tokens.
Let $M=\log_2 (2 \ov{\Lk})=10$.
We train $M-2$ models $\cM_3, \cM_4, \ldots, \cM_M$.
For each $3 \le m \le M$, $\cM_m$ is trained on the dataset $\cD_m$ with context length $\Lc^{(m)} = 2^m$, 
where
    $\cD_m := \{ \cD[i \cdot 2^{m-1}: i \cdot 2^{m-1} + 2^m] \mid i = 0, 1, \ldots \}.$
Note there are overlaps in the sequences of length $2^{m-1} = \Lc^m / 2$.

For any training sequence $s \in \cD_m$ with $\abs{s} = 2^m$, the loss is computed \textbf{\emph{only on the second half}} of the sequence, i.e., treating the first half as hint and the second half as completion:
\begin{align*}
    \cL_m (\theta) = \bbE_{s \sim \cD_m} \left[ \sum_{i=2^{m-1}}^{2^m-1} -\log \cM_\theta (s[i]\ |\ s[:i]) \right].
\end{align*}

The intuition behind this choice is that we want language models to ``think more'' before they ``speak.''
With access to the full context, models can predict future tokens more accurately.
We show ablation results comparing \ding{172} non-overlapping sequences with full loss versus \ding{173} overlapping sequences with loss computed only on the second half in \Cref{tab:main_results}.

In practice, we use different batch sizes when training different models.
Compared to direct training, we set $B_m = 2 B \cdot \Lc / \Lc^{(m)}$, where the coefficient $2$ accounts for the overlapping sequences.
This batch size selection ensures that all models are updated for the same number of steps.

We show that each knowledge occurrence is guaranteed to be captured by some training sequence in \Cref{sec:knowledge_capture}.

\subsubsection{Model ensemble} \label{sec:model_ensemble}

Having trained $M - 2$ models $\cM_3, \ldots, \cM_M$, our next task is to ensemble them to produce valid probability distributions over tokens.
Given an input token array $s$, we predict the next token by first querying each model with its corresponding context window to obtain $M - 2$ probability distributions: for $3 \le m \le M, x \in \Sigma$,
    $p_m (x) := \cM_m (x \mid s[-2^{m-1}:])$.

We then define the confidence of each model by its maximum log probability across the token space: for $3 \le m \le M$,
    $c_m := \max_{x \in \Sigma} \log p_m (x)$.
The weight of each model is calculated by: for $3 \le m \le M$,
\begin{align*}
    w_m &:= \softmax (m \mid \{-\log (- c_{m'})\}_{m'=3}^M )
    = \frac{\exp(-\log (-c_m))}{\sum_{m'=3}^M \exp(-\log (-c_{m'}))} 
    \propto \frac{1}{-c_m}.
\end{align*}
Considering the cases where $c_m$ is extremely close to $0$, the practical implementation is $w_m \propto 1/(\epsilon - c_m)$ where $\epsilon = 10^{-9}$.
The intuition is that we want to emphasize the predictions of models with high certainty while minimizing the influence of less confident models.

Finally, we compute the ensemble model using the weighted mixture of log probabilities as
    $ l (x) := \sum_{m=3}^M w_m \log p_m (x)$, 
for any $x \in \Sigma$.

\para{Evaluation.}
For efficient evaluation, we calculate probabilities for multiple tokens simultaneously with each model.
Specifically, for $3 \le m \le M$, at position $i$, we input the sequence $s[i-2^{m-1}: i+2^{m-1}]$ to model $\cM_m$ to compute logits for positions $i+1, i+2, \ldots, i+2^{m-1}$, then increment $i$ by $2^{m-1}$.
After obtaining logits from each model for all positions, we apply the ensemble method described above to calculate the final probability distribution.

\subsection{Compressing all the models} \label{sec:cascade}

While results in \Cref{tab:main_results} demonstrate the effectiveness of using a series of \emph{\textbf{cascading}} datasets, the increased total model size raises a significant concern.
To address this issue, we compress the models by training a single model $\cM_\theta$ using the average of losses $\{\cL_m\}_{m=3}^M$.
We name this approach as \emph{\textbf{CASCADE}}, which minimizes the \emph{\textbf{CASCADE}} loss defined as:
\begin{align*}
    \cL_{\textup{CASCADE}} (\theta) = \frac{1}{M-2} \sum_{m=3}^M \bbE_{s \sim \cD_m} \left[ \sum_{i=2^{m-1}}^{2^m-1} -\log \cM_\theta (s[i]\ |\ s[:i]) \right].
\end{align*}
During evaluation or inference as described in \Cref{sec:model_ensemble}, we replace all models $\cM_m$ with the single model $\cM_\theta$.
This approach maintains the same model size as the baselines rather than being $8$ times larger.
Theoretically, CASCADE also does not incur higher time complexity as we explained in \Cref{sec:time_complexity}.

\subsection{Results} \label{sec:cascade_results}

\textbf{To ensure fair comparison with dataset rewriting, we evaluate only using the \emph{hold-out} knowledge for \fts\qwiki and \fwiki\qts.}
For a comprehensive analysis, we implemented both training configurations: non-overlapping training sequences with loss computed on the full sequence, and overlapping training sequences with loss computed only on the second half of each sequence.

\para{Ablation on practical running time.}
In practice, the running time of a forward pass can be reduced significantly to an almost linear dependence on sequence length using FlashAttention \citep{dao2022flashattention,dao2023flashattention}.
When training for the same number of epochs, the CASCADE loss requires approximately $M-2$ times the training time of the baseline method (direct training).
For a fair comparison, we conducted an ablation study allowing the baseline method (with context length $1024$) to train for the same duration as CASCADE.

We present the normalized log probabilities of (original) \emph{\textbf{CASCADE}} and ablation studies in \Cref{tab:main_results}.

\begin{table}[htbp]
    \centering
    \resizebox{0.99\linewidth}{!}{%
    \begin{tabular}{cc!{\vrule width 1.25pt}c|c!{\vrule width 1pt}c|c}
        \noalign{\hrule height 1.5pt}
        \multicolumn{2}{c!{\vrule width 1.25pt}}{Methods} & \fts\qts & \fwiki\qwiki & \fts\qwiki & \fwiki\qts \\
        \noalign{\hrule height 1.25pt}
        \multirow{2}{*}{\scriptsize \makecell{Direct\\Training\\(Ablation)}} & Non-overlap & $-1.93 \times 10^{-8}$ & $-1.43 \times 10^{-8}$ & $-4.77 \times 10^{-3}$ & $-1.53 \times 10^{-2}$ \\
        \cline{2-6}
        & Overlap & $-2.29 \times 10^{-8}$ & $-2.16 \times 10^{-7}$ & $-2.66 \times 10^{-1}$ & $-4.31 \times 10^{-1}$ \\
        \noalign{\hrule height 1pt}
        \multirow{2}{*}{\makecell{Original\\CASCADE}} & Non-overlap & $-5.91 \times 10^{-6}$  & $-6.21 \times 10^{-6}$  & $-2.45 \times 10^{-5}$  & $-1.36 \times 10^{-4}$ \\
        \cline{2-6}
        & Overlap & $-9.65 \times 10^{-9}$  & $-8.51 \times 10^{-9}$  & $\red{\boldsymbol{-2.59 \times 10^{-8}}}$  & $\red{\boldsymbol{-9.22 \times 10^{-7}}}$ \\
        \noalign{\hrule height 1pt}
        \multirow{2}{*}{CASCADE} & Non-overlap & $-3.87 \times 10^{-5}$  & $-3.95 \times 10^{-5}$  & $-1.87 \times 10^{-4}$  & $-1.54 \times 10^{-4}$ \\
        \cline{2-6}
        & Overlap & $-3.26 \times 10^{-7}$  & $-3.44 \times 10^{-7}$  & $\red{\boldsymbol{-3.71 \times 10^{-6}}}$  & $\red{\boldsymbol{-5.06 \times 10^{-6}}}$ \\
        \noalign{\hrule height 1.5pt}
    \end{tabular}
    }
    \caption{Normalized log probabilities for different methods, averaged over $5$ random seeds. Cross-mode results better than or equal to the order of $10^{-6}$ are in bolt red text.}
    \label{tab:main_results}
\end{table}

More results can be found in \Cref{sec:quantitative_results}:
To better illustrate the contribution of different context lengths during evaluation, we display the normalized log probabilities when using only a single context length (\emph{without model ensemble}) in \Cref{tab:cascade_single_ctx_len}, with specific context lengths \emph{excluded} in \Cref{tab:cascade_without_ctx_len}, and visualize the weight vector $\{w_m\}_{3 \le m \le M}$ at each token position for various knowledge lengths in \Cref{fig:weights}.
To show the generalization capability of CASCADE, we compare it with direct training and dataset rewriting when there are $3$ modes (\Cref{tab:3_modes}).
We also show the performance of baselines when using a roughly double-sized model in \Cref{tab:3_modes_bigger_model}.



\subsection{Remarks}

We now make several remarks for CASCADE.

$\bullet$ \textbf{\emph{Cascading} the dataset substantially enhances the cross-mode knowledge retrieval capability of language models.}
Results in \Cref{tab:main_results} demonstrate that with \emph{\textbf{cascading}} datasets, models achieve significantly improved cross-mode knowledge retrieval with overlapping sequences compared to non-overlapping sequences, and most importantly, outperform all baselines presented in \Cref{sec:ratio_results}.
As anticipated, model compression introduces a minor performance degradation.

$\bullet$ \textbf{$\cL_{\textup{CASCADE}}$ functions as an implicit regularizer.}
Unexpectedly, \Cref{tab:cascade_single_ctx_len} shows that training with $\cL_{\textup{CASCADE}}$ alone, even without model ensemble, improves cross-mode knowledge retrieval capability.
This loss function appears to implicitly regularize the language model against spurious correlations.
Comparing \Cref{tab:cascade_single_ctx_len} and the rows corresponding to CASCADE in \Cref{tab:main_results}, we observe that model ensemble further enhances performance by an order of magnitude.

$\bullet$ \textbf{Small context lengths are critical for initial positions.}
\Cref{fig:weights} illustrates that for the first few tokens in the completion, models with smaller context lengths exhibit greater prediction certainty.
Additionally, the context lengths of $128$ and $256$ appear to be excessive according to \Cref{tab:cascade_without_ctx_len}.

$\bullet$ \textbf{CASCADE delivers benefits beyond simply increasing training epochs.}
Rows corresponding to direct training in \Cref{tab:main_results} confirm that merely extending training time does not enable baselines to match CASCADE's performance, with results on \fts\qwiki and \fwiki\qts remaining significantly inferior to CASCADE.
Furthermore, in this ablation study, non-overlapping sequences notably outperform overlapping sequences.
This occurs because cascading context lengths are essential for capturing ``local'' information, whereas using a single large context length and calculating loss only on the second half disrupts these ``local'' connections.

$\bullet$ \textbf{CASCADE generalizes to more modes better than dataset rewriting.}
\Cref{tab:3_modes} confirms CASCADE's advantage over direct training and full rewriting when there are $3$ modes.
The row corresponding to full rewriting without one mode-knowledge pair corroborates the remark in \Cref{sec:rewrite_remark} that omitting a mode-knowledge pair directly harms its evaluation performance, demonstrating that meaningful rewriting requires coverage of all pairs.

$\bullet$ \textbf{CASCADE's performance matches that of full rewriting while using less than half the model size.}
Comparing \Cref{tab:3_modes,tab:3_modes_bigger_model}, increasing model size from 162M to 350M improves each method's performance by approximately one order of magnitude.
With this increase, full rewriting (350M) matches CASCADE (162M), while direct training still struggles with cross-mode knowledge retrieval.
\section{Conclusion}

We investigated language models' cross-mode knowledge retrieval capability from both qualitative and quantitative perspectives.
Our qualitative pipeline reveals that LLMs such as GPT-4o cannot perform cross-mode knowledge retrieval satisfactorily.
Quantitatively, we formulated this problem using two format datasets as modes and random token sequences as knowledge, and experimented with a straightforward approach -- dataset rewriting -- showing that only substantial dataset rewriting efforts can alleviate this issue.
Finally, we proposed CASCADE, a novel pretraining method, along with its model-compression version.
Experiments demonstrate that CASCADE significantly outperforms baselines.

Despite its fundamental nature, our work has several limitations that may inspire future studies.
First, we did not apply our training method to real-world datasets due to limited computational resources and lack of evaluation metrics.
The qualitative pipeline in \Cref{sec:qualitative_gen} may serve as a metric, but automatically selecting representative knowledge merits further study.
Second, our study contains at most $3$ modes.
Future work could study an $n$-mode setting where $n > 3$ and compute the corresponding normalized log probabilities to verify CASCADE's advantage over dataset rewriting.
Also, researchers can aim at finding other novel ideas to get better performances.

\section*{Acknowledgement}
RZ acknowledges helpful discussions with and support of (ordered lexicographically) Harkirat Behl, Sebastien Bubeck, Tong Chen, Yifang Chen, Simon Du, Beibin Li, Yuanzhi Li, Chen Liang, Zhihan Xiong, Dingli Yu, Cyril Zhang, and Xinran Zhao.

\bibliography{ref}
\bibliographystyle{COLM2025/colm2025_conference}

\appendix

\newpage
\section{Additional related works} \label{sec:additional_related_works}

\para{Physics of language models.}
A line of closely related works are physics of language models \citep{allenzhu2024physicslanguagemodels1,ye2024physicslanguagemodels21,ye2024physicslanguagemodels22,allenzhu2024physicslanguagemodels31,allenzhu2024physicslanguagemodels32,allenzhu2024physicslanguagemodels33}, which center around how language models learn and manipulate knowledge.
\cite{allenzhu2024physicslanguagemodels1} demonstrates that transformer-based models like GPT \citep{radford2018improving,radford2019language,brown2020language,achiam2023gpt} can effectively learn and generate complex, recursive language structures from context-free grammars.
In \cite{ye2024physicslanguagemodels21}, the authors investigate how small language models solve grade-school math problems, distinguishing between memorization and genuine reasoning.
\cite{ye2024physicslanguagemodels22} focuses on improving models' reasoning accuracy by incorporating``retry data'' during pretraining stage.
\cite{allenzhu2024physicslanguagemodels31} finds that knowledge augmentation during pretraining significantly improves the models' ability to extract and utilize knowledge, introduces novel probing techniques to understand this process, and suggests to enhance language model training with \emph{data rewriting} and early introduction of question-answering tasks.
\cite{allenzhu2024physicslanguagemodels32} explores the limitations of language models in executing basic knowledge manipulation tasks—retrieval, classification, comparison, and inverse search.
It proposes methods like generating more Chain-of-Though (CoT, \cite{wei2022chain}) data, employing retrieval augmented generation (RAG, \cite{lewis2020retrieval}) and reversal training.
\cite{allenzhu2024physicslanguagemodels33} presents a comprehensive study on the knowledge capacity scaling laws of language models, revealing that a $2$bit/param capacity ratio is achievable across various architectures and training conditions, but is affected by factors such as training exposure, model architecture, quantization, sparsity, and the quality of training data.

\para{Spurious correlations.}
Spurious correlations represent a significant threat to the reliability and trustworthiness of NLP systems, as they can cause models to learn unintended shortcuts rather than the underlying task-relevant signals \citep{eisenstein-2022-informativeness,51325}.
This issue has been widely studied in text classifications tasks.
\citet{joshi2022all} examines spurious features through a causal lens, classifying them based on probability of necessity (PN) and probability of sufficiency (PS).
They identify two categories: irrelevant features (low PN, low PS) and necessary features (high PN, low PS).
\citet{wu2022generating} introduce a data generation approach to mitigate spurious correlations by creating debiased versions of datasets.
\citet{bansal2023controlling} estimate the causal effect of features on labels and regularize models to match this true effect, developing an automated augmentation method that improves performance on minority groups while maintaining overall accuracy.
\citet{lee2024clarify} build a human-model interaction interface, allowing users to give descriptions about models' misconceptions about spurious correlations, ultimately improving the performance.

\section{Qualitative studies} \label{sec:qualitative_gen}

\subsection{Setup}

For qualitative studies, we use paragraphs from Wikipedia as test cases.
We manually select a sentence from the original Wikipedia text, replace it with a \texttt{[BLANK]} along with its hint.
We call this input \texttt{original}, and the selected sentence is called \texttt{answer}.

Next, we prompt GPT-4o using the template in \Cref{text:rewrite_template}, replacing \texttt{\{text\}} with \texttt{original}.
This generates a story-style text called \texttt{altered}, which contains a corresponding \texttt{[BLANK]} with a hint.

\begin{textbox}[htbp]
\begin{textframe}
\begin{lstlisting}[language=text]
You will help me rewrite a text into another style.
I will give you a text based on a fact from Wikipedia.
I left a blank, [BLANK], as well as its hint in the text.
Your task is to rewrite the text into a story, under the setting that a mother is telling a bedtime story to her kid.
Aside from the information in the original text, you should describe about the environment, the characters, and the plot.
The rewritten text should be coherent and consistent with the original text.
You must retain the blank and its hint in the rewritten text, for example, when the hint requires to output three items, you should include the hint in the rewritten text as well.

===== Text =====
{text}
\end{lstlisting}
\end{textframe}
\caption{The input template for rewriting into a story style.}
\label{text:rewrite_template}
\end{textbox}

We then separately prompt GPT-4o $100$ times using the template in \Cref{text:completion_template}, replacing \texttt{\{text\}} with \texttt{original} and \texttt{altered}, respectively.
To avoid API-side caching, we add \texttt{ATTEMPT \{i\}} to the beginning of each prompt.
This generates responses $r_{\texttt{original}}^{(i)}$ and $r_{\texttt{altered}}^{(i)}$ for $1 \le i \le 100$.

\begin{textbox}[htbp]
\begin{textframe}
\begin{lstlisting}[language=text]
I will give you a text based on a fact.
I left a blank, [BLANK], as well as its hint in the text.
Please fill in the blank after you read the text.
You should provide the most appropriate information, as accurate as possible.

===== Text =====
{text}
\end{lstlisting}
\end{textframe}
\caption{The input template for blank completion.}
\label{text:completion_template}
\end{textbox}

Finally, we prompt GPT-4o using the template in \Cref{text:judge_template}, replacing \texttt{\{text\}} with \texttt{type}, \texttt{\{response\}} with $r_{\texttt{{type}}}^{(i)}$, and \texttt{\{answer\}} with \texttt{answer}, where $\texttt{type} \in \{ \texttt{original}, \texttt{altered} \}$ and $1 \le i \le 100$.
We extract accuracies from the judge outputs and average them.

\begin{textbox}[htbp]
\begin{textframe}
\begin{lstlisting}[language=text]
You are a judge to evaluate the response of the completion system.
I'll provide you a text with a blank, [BLANK].
Then, I'll provide you a response to fill in the blank, and its ground truth answer.
Please evaluate whether the response is correct or not, output a float number between 0 and 1 to represent the accuracy.
Identify each important aspects in the ground truth answer, and compare them with the response.
The floating number should be finally outputed in the following format:
```Accuracy
[ACCURACY]
```

===== Text =====
{text}

===== Response =====
{response}

===== Ground Truth =====
{answer}
\end{lstlisting}
\end{textframe}
\caption{The input template for judging a completion response.}
\label{text:judge_template}
\end{textbox}

\subsection{Results}

We present three examples in \Cref{tab:gen1,tab:gen2,tab:gen3}.
Detailed results are included in \texttt{scripts/eval/data.json} in the supplementary materials.

\begin{table*}[htbp]
\footnotesize
    \centering
        \centering
        \small
        \scalebox{.8}{
\begin{tabular}{p{1.5cm}|p{15cm}}
\toprule[1.2pt]
\texttt{original} & \lstinputlisting[language=text]{original1}\\
\hline
\texttt{answer} & \lstinputlisting[language=text]{answer1} \\
\hline
Example response & \lstinputlisting[language=text]{original1output} \\
\hline
Judge & \lstinputlisting[language=text]{original1outputverdict} \\
\hhline{==}\\
\texttt{altered} & \lstinputlisting[language=text]{altered1}\\
\hline
Example response & \lstinputlisting[language=text]{altered1output} \\
\hline
Judge & \lstinputlisting[language=text]{altered1outputverdict} \\
\bottomrule[1.2pt]
\end{tabular}
    }
\caption{Example 1: the average accuracies of the responses for the original input and altered input are $48.0\%$ and $25.9\%$, respectively.}
\label{tab:gen1}
\end{table*}

\begin{table*}[htbp]
\footnotesize
    \centering
        \centering
        \small
        \scalebox{.8}{
\begin{tabular}{p{1.5cm}|p{15cm}}
\toprule[1.2pt]
\texttt{original} & \lstinputlisting[language=text]{original2}\\
\hline
\texttt{answer} & \lstinputlisting[language=text]{answer2} \\
\hline
Example response & \lstinputlisting[language=text]{original2output} \\
\hline
Judge & \lstinputlisting[language=text]{original2outputverdict} \\
\hhline{==}\\
\texttt{altered} & \lstinputlisting[language=text]{altered2}\\
\hline
Example response & \lstinputlisting[language=text]{altered2output} \\
\hline
Judge & \lstinputlisting[language=text]{altered2outputverdict} \\
\bottomrule[1.2pt]
\end{tabular}
    }
\caption{Example 2: the average accuracies of the responses for the original input and altered input are $93.3\%$ and $62.0\%$, respectively.}
\label{tab:gen2}
\end{table*}

\begin{table*}[htbp]
\footnotesize
    \centering
        \centering
        \small
        \scalebox{.8}{
\begin{tabular}{p{1.5cm}|p{15cm}}
\toprule[1.2pt]
\texttt{original} & \lstinputlisting[language=text]{original3}\\
\hline
\texttt{answer} & \lstinputlisting[language=text]{answer3} \\
\hline
Example response & \lstinputlisting[language=text]{original3output} \\
\hline
Judge & \lstinputlisting[language=text]{original3outputverdict} \\
\hhline{==}\\
\texttt{altered} & \lstinputlisting[language=text]{altered3}\\
\hline
Example response & \lstinputlisting[language=text]{altered3output} \\
\hline
Judge & \lstinputlisting[language=text]{altered3outputverdict} \\
\bottomrule[1.2pt]
\end{tabular}
    }
\caption{Example 3: the average accuracies of the responses for the original input and altered input are $78.3\%$ and $28.5\%$, respectively.}
\label{tab:gen3}
\end{table*}

\newpage
\section{Justifications for CASCADE}

\subsection{Explanation for knowledge capture} \label{sec:knowledge_capture}

Here we justify that this cascading design of datasets ensures that each piece of knowledge is captured by \emph{at least one} language model.
Consider a piece of knowledge with length $\Lk$ that appears in position $(p, p+1, \ldots, p+\Lk-1)$ of $\cD$.
Then for each $3 \le m \le M$, we identify the training sequences which contain the knowledge across the halfway point (i.e., sequences that have this knowledge in both the hint and completion parts):
\begin{align*}
    \left\{\begin{array}{ll}
        i \cdot 2^{m-1} \le p, &\textup{\ding{172} Training sequence starts before \knw;} \\
        i \cdot 2^{m-1} + 2^{m-1} > p, &\textup{\ding{173} Hint contains \knw;} \\
        i \cdot 2^{m-1} + 2^{m-1} \le p + \Lk - 1, &\textup{\ding{174} Completion contains \knw}.
    \end{array}\right.
\end{align*}
With all requirements combined, we can solve for $i$:
\begin{align}
    \frac{p}{2^{m-1}} - 1 < i \le \min \left\{ \frac{p}{2^{m-1}}, \frac{p+\Lk-1}{2^{m-1}} - 1 \right\}. \label{eq:sol_all_req}
\end{align}
If $\Lk \ge 2^{m-1}+1 > \Lc^m / 2$, then there is a unique solution $i = \floor{p/2^{m-1}}$.
Here requirement \ding{172} is optional, because without it means the training sequence does not have mode in the hint part, which is helpful for knowledge completion.
Thus, for all $m \le 1 + \floor{\log (\Lk-1)}$, this piece of knowledge occupies half of a training sequence in $\cD_m$ and is therefore captured by model $\cM_m$.

\subsection{Theoretical time complexity analysis for CASCADE} \label{sec:time_complexity}

In self-attention \citep{waswani2017attention}, processing a batch of $B$ training sequences with length $\Lc$ takes $\Theta (B (\Lc)^2)$ time.

\para{Training/Evaluation.}
Suppose we use the efficient evaluation method in \Cref{sec:model_ensemble}, then training and evaluation are essentially the same (except for a backward pass).
The time complexity is
\begin{align*}
    \sum_{m=3}^M \Theta ( B_m (\Lc^{(m)})^2 ) = \sum_{m=3}^M \Theta ( 2 B \Lc \Lc^{(m)} ) = \sum_{m=3}^M \Theta ( 2 B \Lc 2^m ) = \Theta ( B (\Lc)^2 )
\end{align*}
as we recall $M = \log_2 (2 \ov{\Lk}) = \log_2 \Lc$.

\para{Inference.}
Suppose batch size $B = 1$ in inference.
To generate a single sequence using the original method, the time complexity is
\begin{align*}
    \sum_{p=1}^{\Lc} \Theta ( p^2 ) = \Theta ( (\Lc)^3 ).
\end{align*}
For CASCADE, the time complexity is
\begin{align*}
    \sum_{p=1}^{\Lc} \sum_{m=3}^M \Theta (\min\{p^2, (\Lc^{(m)}/2)^2\}) \le \sum_{p=1}^{\Lc} \sum_{m=3}^M \Theta ((\Lc^{(m)}/2)^2) = \sum_{p=1}^{\Lc} \sum_{m=3}^M \Theta (4^m) = \Theta ((\Lc)^3)
\end{align*}
as we recall $M = \log_2 \Lc$.

Therefore, from a theoretical perspective, CASCADE does not introduce much time overhead.

\newpage
\section{Experiment details for the quantitative experiments} \label{sec:quantitative_exp}

\subsection{Datasets}

There are $473992236$ tokens in \Fts, $484159419$ tokens in \Fwiki, and $484626190$ tokens in \Fcode (mentioned in \Cref{sec:3_modes}).
When constructing \Dts, \Dwiki and \Dcode, regardless of the random seed, the data are arranged in a fixed order such that all types (\fwiki\kwiki, \fwiki\kts, \fwiki\kcode, \fts\kts, \fts\kwiki, \fts\kcode, \fcode\kcode, \fcode\kwiki, \fcode\kts) of data are approximately uniformly distributed.
All the datasets use the same \emph{dataset seed} $42$ which is independent of the random seeds in training, to ensure that all the datasets are the same across different runs.
The hyperparameters for dataset construction are listed in \Cref{tab:dataset_hparam}.

\begin{table*}[ht]
    \centering
    
    \begin{tabular}{l | l}
    \hline
    \textbf{Hyperparameter} & \textbf{Value}  \\
    \hline
    Random token sequence & \\
    - Length range & $[8, 512]$ \\
    - Different sequences per dataset & $32$ \\
    - $\Nocc$ & $8192$ \\
    - $\Noccx$ & $\{0\} \cup \{2^i \mid 1 \le i \le 13\}$\\
    \hline 
    
    Dataset seed & $\{42\}$ \\
    \hline
    \end{tabular}
    \caption{Hyperparameters of datasets}
    \label{tab:dataset_hparam}
\end{table*}

\subsection{Models} \label{sec:model_config}

We use a customized phi-1 (162M) model specified in \Cref{tab:model_spec}, where the original architecture is introduced in \citet{gunasekar2023textbooks}.
The value of n\_positions corresponds to the sequence lengths in training (see \Cref{tab:hparam}).

\begin{table*}[ht]
    \centering
    
    \begin{tabular}{l | l}
        \hline
        \textbf{Specification} & \textbf{Value}  \\
        \hline
        Type & mixformer-sequential \\
        \hline
        Architecture & \\
        - block\_cls & parallel \\
        - mixer & \\
        \ \ - mixer\_cls & mha \\
        \ \ - dropout & $0.1$ \\
        - mlp\_cls & fused\_mlp \\
        \hline
        Total parameters & $162$m \\
        - vocab\_size & $50304$ \\
        - n\_positions & $\{2^m\mid 3 \le m \le 10\}$ \\
        - n\_embd & $768$ \\
        - n\_layer & $12$ \\
        - n\_head & $12$ \\
        - rotary\_dim & $32$ \\
        \hline
        resid\_pdrop & $0.1$ \\
        \hline
    \end{tabular}
    \caption{Model specification of our customized phi-1 (162M).}
    \label{tab:model_spec}
\end{table*}

\subsection{Training} \label{sec:exp_ratio_details}

We use the set of hyperparameters in \Cref{tab:hparam}.
For the experiments of \emph{ablation on practical running time} in \Cref{sec:cascade_results}, we use $16$ epochs, while for all the other experiments, we use $2$ epochs.
For the experiments for dataset rewriting in \Cref{sec:ratio_results}, we use sequence lengths of $1024$, while for the experiments of cascading datasets in \Cref{sec:cascade_results}, $m$ can vary between $3,4, \ldots, 10$.

\begin{table*}[ht]
    \centering
    
    \begin{tabular}{l | l}
        \hline
        \textbf{Hyperparameter} & \textbf{Value}  \\
        \hline
        Number of epochs & $\{2, 16\}$ \\
        \hline
        Train batch size & $1024$ \\
        \hline
        Optimizer & AdamW \\
        - Gradient clipping norm & $1.0$ \\
        - $\beta_1, \beta_2$ & $0.9, 0.95$ \\
        - $\epsilon$ & $1 \times 10^{-7}$ \\
        - Weight decay & $0.1$ \\
        \hline
        Learning rate scheduler & WarmupDecayLR \\
        - Warmup min lr & $1 \times 10^{-7}$ \\
        - Warmup max lr & $1 \times 10^{-4}$ \\
        - Warmup steps & $500$ \\
        - Warmup type & Linear \\
        \hline
        Precision & fp$16$ (initial scale power: $12$)\\
        \hline
        Sequence length & $\{2^m \mid 3 \le m \le 10\}$ \\
        \hline
        Random seed & $\{42, 142857, 2225393, 20000308, 2018011309\}$ \\
        \hline
    \end{tabular}
    \caption{Hyperparameters of the quantitative experiments}
    \label{tab:hparam}
\end{table*}

\subsection{More results} \label{sec:quantitative_results}

Here we list results not being able to be presented in the main text due to page limit.

\subsubsection{Evaluating each context length in CASCADE}

\begin{figure}[htbp]
    \centering
    \includegraphics[width=0.95\linewidth]{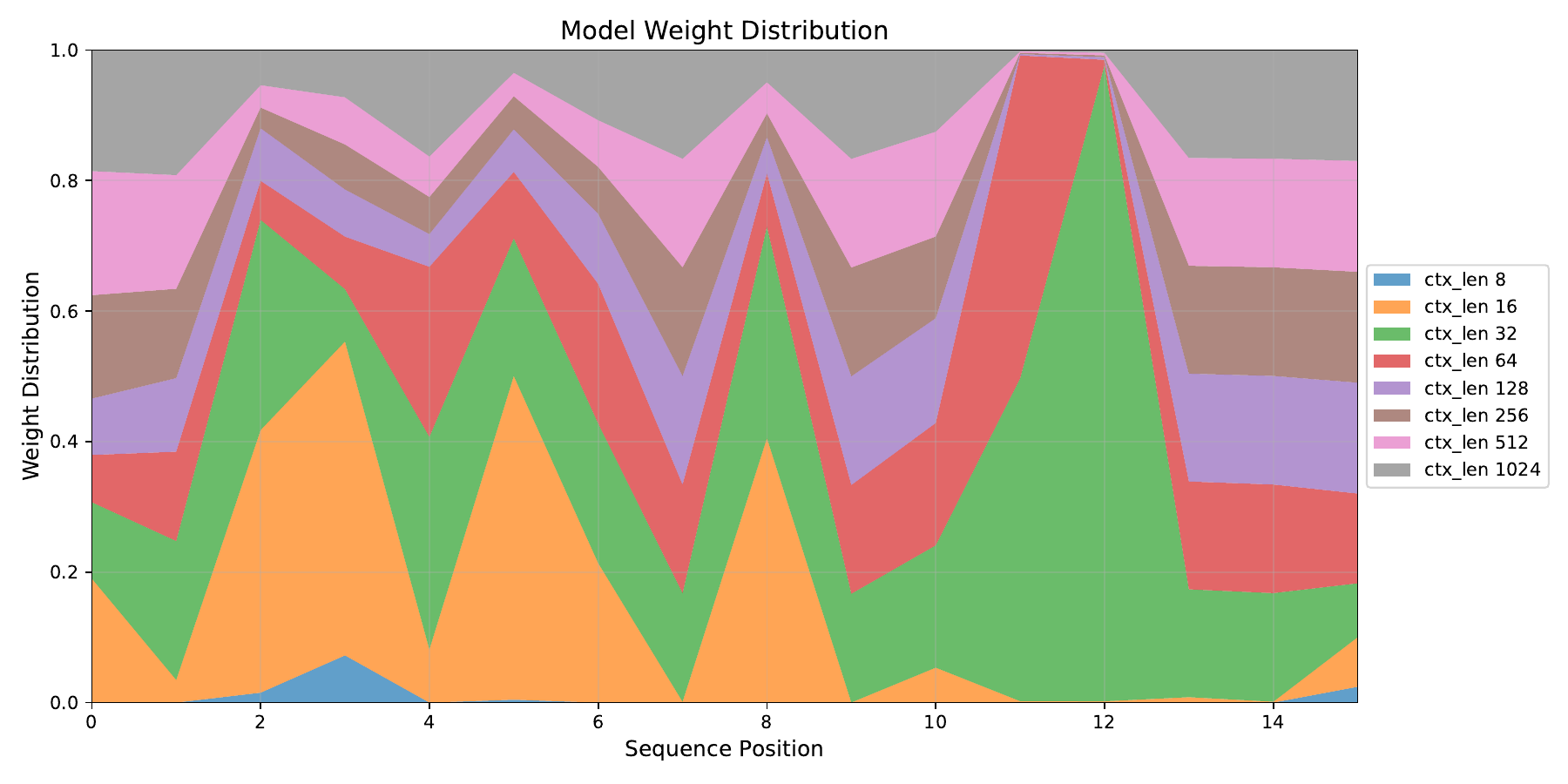}
    \includegraphics[width=0.95\linewidth]{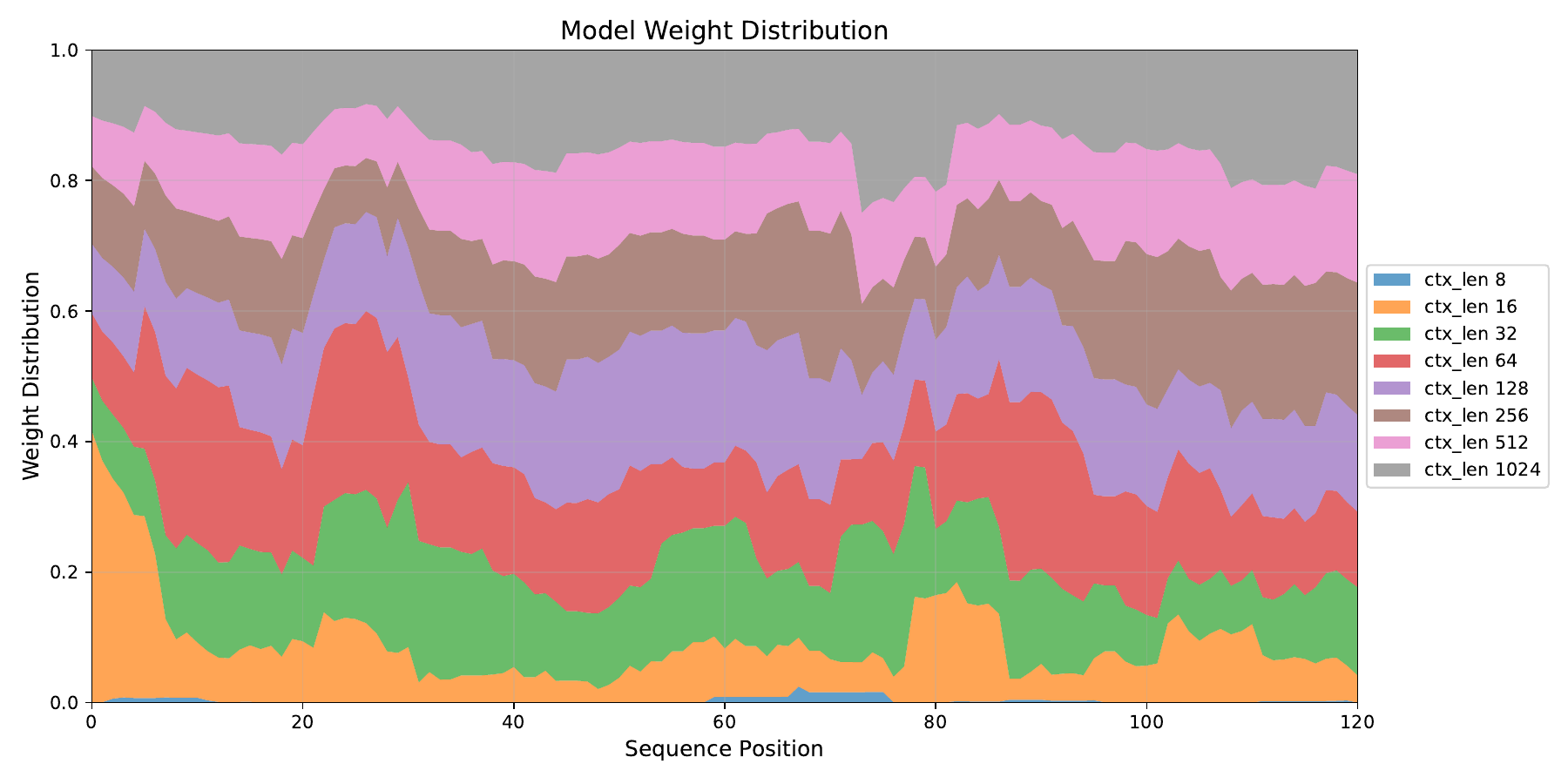}
    \includegraphics[width=0.95\linewidth]{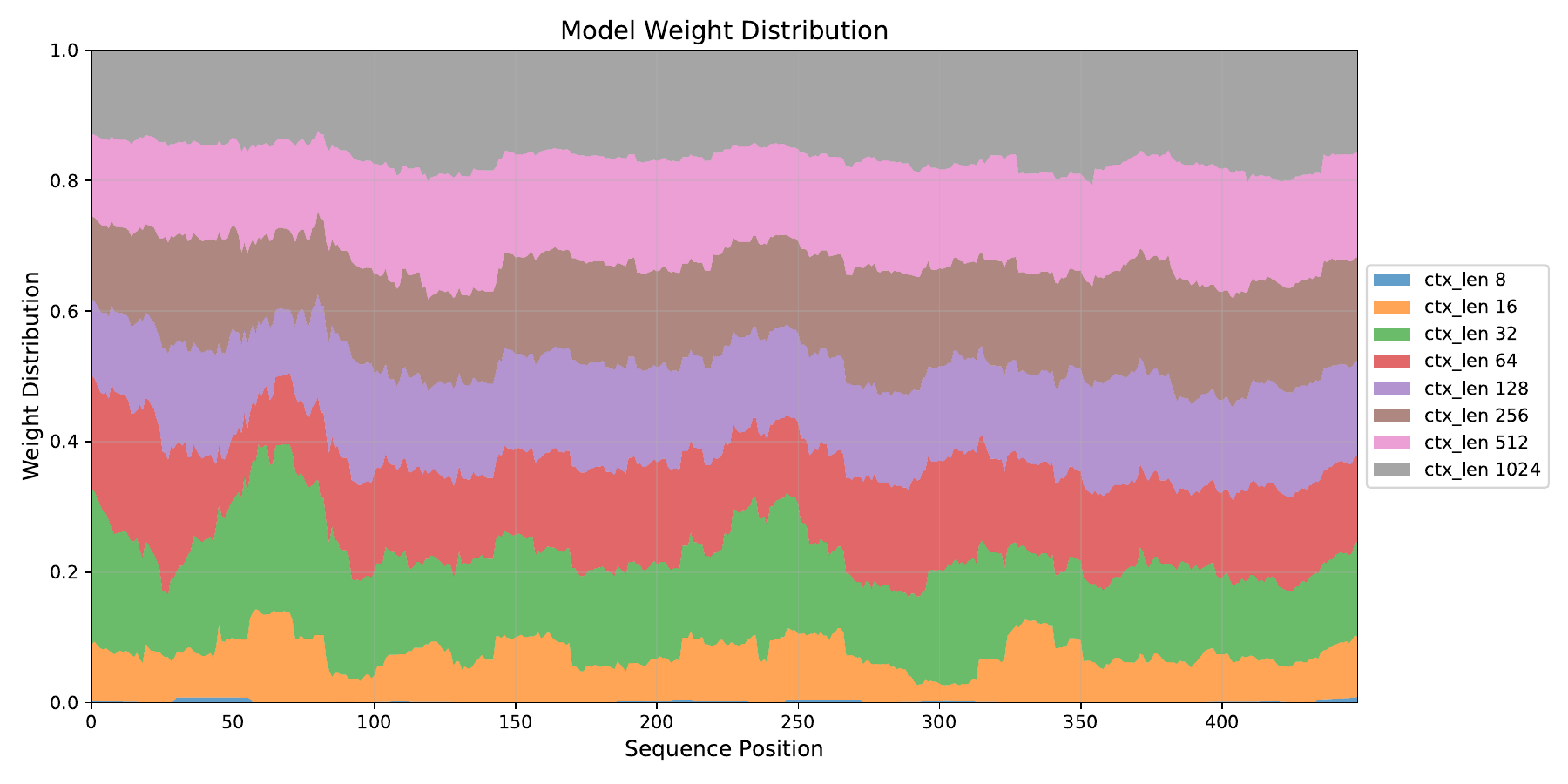}
    \caption{Weight distribution over models for different positions in the completion part.}
    \label{fig:weights}
\end{figure}

To see how each context length $\Lc^{(m)}$ contributes to the performance of $\cM_\theta$, we conducted $3$ extra evaluations: (1) Evaluation with exactly one context length (\Cref{tab:cascade_single_ctx_len}), (2) Evaluation without exactly one context length (\Cref{tab:cascade_without_ctx_len}), and (3) Visualization of the contributions of each context length at each sequence position (\Cref{fig:weights}).

\begin{table}[htbp]
    \centering
    \begin{tabular}{c!{\vrule width 1.25pt}c|c!{\vrule width 1pt}c|c}
        \noalign{\hrule height 1.5pt}
        Context Length & \fts\qts & \fwiki\qwiki & \fts\qwiki & \fwiki\qts \\
        \noalign{\hrule height 1.25pt}
        $ 8 $  & $-5.00 \times 10^{-1}$  & $-4.98 \times 10^{-1}$  & $-4.98 \times 10^{-1}$  & $-5.00 \times 10^{-1}$ \\
        \hline
        $ 16 $  & $-3.21 \times 10^{-5}$  & $-2.64 \times 10^{-5}$  & $-4.11 \times 10^{-5}$  & $-5.56 \times 10^{-5}$ \\
        \hline
        $ 32 $  & $-6.74 \times 10^{-7}$  & $-6.65 \times 10^{-7}$  & $-1.01 \times 10^{-5}$  & $-1.82 \times 10^{-5}$ \\
        \hline
        $ 64 $  & $-4.94 \times 10^{-7}$  & $-5.08 \times 10^{-7}$  & $-1.35 \times 10^{-5}$  & $-1.72 \times 10^{-5}$ \\
        \hline
        $ 128 $  & $-4.35 \times 10^{-7}$  & $-4.63 \times 10^{-7}$  & $-1.27 \times 10^{-5}$  & $-1.78 \times 10^{-5}$ \\
        \hline
        $ 256 $  & $-4.07 \times 10^{-7}$  & $-4.91 \times 10^{-7}$  & $-1.39 \times 10^{-5}$  & $-1.55 \times 10^{-5}$ \\
        \hline
        $ 512 $  & $-3.85 \times 10^{-7}$  & $-5.33 \times 10^{-7}$  & $-1.68 \times 10^{-5}$  & $-1.62 \times 10^{-5}$ \\
        \hline
        $ 1024 $  & $-3.68 \times 10^{-7}$  & $-5.58 \times 10^{-7}$  & $-2.00 \times 10^{-5}$  & $-2.17 \times 10^{-5}$ \\
        \noalign{\hrule height 1.5pt}
    \end{tabular}
    \caption{Normalized log probabilities of the model trained using CASCADE with overlapping sequences, evaluated using individual fixed context lengths. The values are averaged over $5$ random seeds.}
    \label{tab:cascade_single_ctx_len}
\end{table}

\begin{table}[htbp]
    \centering
    \begin{tabular}{c!{\vrule width 1.25pt}c|c!{\vrule width 1pt}c|c}
        \noalign{\hrule height 1.5pt}
        Context Length & \fts\qts & \fwiki\qwiki & \fts\qwiki & \fwiki\qts \\
        \noalign{\hrule height 1.25pt}
$ 8 $  & $ -3.27 \times 10^{-7} $  & $ -3.46 \times 10^{-7} $  & $ -3.71 \times 10^{-6} $  & $ \red{\boldsymbol{ -5.05 \times 10^{-6} }} $ \\
\hline
$ 16 $  & $ -3.42 \times 10^{-7} $  & $ -3.68 \times 10^{-7} $  & $ -5.37 \times 10^{-6} $  & $ -8.30 \times 10^{-6} $ \\
\hline
$ 32 $  & $ -3.35 \times 10^{-7} $  & $ -3.60 \times 10^{-7} $  & $ -4.08 \times 10^{-6} $  & $ -5.59 \times 10^{-6} $ \\
\hline
$ 64 $  & $ \red{\boldsymbol{ -3.23 \times 10^{-7} }} $  & $ -3.45 \times 10^{-7} $  & $ -3.85 \times 10^{-6} $  & $ \red{\boldsymbol{ -4.95 \times 10^{-6} }} $ \\
\hline
$ 128 $  & $ \red{\boldsymbol{ -3.22 \times 10^{-7} }} $  & $ \red{\boldsymbol{ -3.43 \times 10^{-7} }} $  & $ \red{\boldsymbol{ -3.66 \times 10^{-6} }} $  & $ \red{\boldsymbol{ -4.89 \times 10^{-6} }} $ \\
\hline
$ 256 $  & $ \red{\boldsymbol{ -3.24 \times 10^{-7} }} $  & $ \red{\boldsymbol{ -3.42 \times 10^{-7} }} $  & $ \red{\boldsymbol{ -3.68 \times 10^{-6} }} $  & $ \red{\boldsymbol{ -5.04 \times 10^{-6} }} $ \\
\hline
$ 512 $  & $ -3.28 \times 10^{-7} $  & $ \red{\boldsymbol{ -3.44 \times 10^{-7} }} $  & $ \red{\boldsymbol{ -3.67 \times 10^{-6} }} $  & $ \red{\boldsymbol{ -5.02 \times 10^{-6} }} $ \\
\hline
$ 1024 $  & $ -3.37 \times 10^{-7} $  & $ -3.50 \times 10^{-7} $  & $ \red{\boldsymbol{ -3.50 \times 10^{-6} }} $  & $ -5.12 \times 10^{-6} $ \\
\noalign{\hrule height 1.5pt}
    \end{tabular}
    \caption{Normalized log probabilities of the model trained using CASCADE with overlapping sequences, evaluated \emph{without} specific context lengths. The values are averaged over $5$ random seeds.
    Values better than those in \Cref{tab:main_results} (CASCADE Overlap) are in bold red text.}
    \label{tab:cascade_without_ctx_len}
\end{table}

\subsubsection{Extension to $3$ modes and bigger models} \label{sec:3_modes}

To show the generalization capability of CASCADE, we conducted additional experiments on $3$ modes.
We used Python code from the \texttt{bigcode/the-stack} dataset \citep{Kocetkov2022TheStack} as a new mode --code.
In \Cref{tab:3_modes}, we present results for dataset rewriting with $r = \infty$ (direct training), $r = 1$ (full rewriting), $r = 1$ but without \fwiki \qts in the training set, and CASCADE using the customized phi-1 (162M) specified in \Cref{sec:model_config}.
Further, we use the model architecture of phi-1-small (350M) \citep{gunasekar2023textbooks}, which roughly doubles the size of our customized phi-1 (162M), to see how baselines scale.
The results are shown in \Cref{tab:3_modes_bigger_model}.

\newpage
\begin{landscape}

\vspace*{2cm}

\begin{table}[ht]
    \centering
    \resizebox{0.99\linewidth}{!}{%
    \begin{tabular}{c!{\vrule width 1.25pt}c|c|c!{\vrule width 1pt}c|c|c|c|c|c}
        \noalign{\hrule height 1.5pt}
        & \fts\qts & \fwiki\qwiki & \fcode\qcode & \fts\qwiki & \fts\qcode & \fwiki\qts & \fwiki\qcode & \fcode\qts & \fcode\qwiki \\
        \noalign{\hrule height 1.25pt}
        Direct training ($r = \infty$) & $-7.74 \times 10^{-6}$ & $-8.39 \times 10^{-6}$ & $-1.75 \times 10^{-5}$ & $-0.215$ & $-0.755$ & $-0.368$ & $-8.79 \times 10^{-2}$ & $-0.724$ & $-2.80 \times 10^{-2}$ \\ \hline
        Full rewriting ($r = 1$) & $-2.62 \times 10^{-5}$ & $-2.78 \times 10^{-5}$ & $-3.10 \times 10^{-5}$ & $-6.89 \times 10^{-5}$ & $-1.33 \times 10^{-4}$ & $-3.46 \times 10^{-4}$ & $-8.45 \times 10^{-5}$ & $-1.51 \times 10^{-4}$ & $-1.84 \times 10^{-4}$ \\ \hline
        $r = 1$ without \fwiki\qts & $-2.87 \times 10^{-5}$ & $-6.37 \times 10^{-6}$ & $-2.52 \times 10^{-5}$ & $-2.02 \times 10^{-4}$ & $-7.71 \times 10^{-5}$ & $-4.67 \times 10^{-3}$ (*) & $-1.37 \times 10^{-4}$ & $-1.52 \times 10^{-4}$ & $-6.32 \times 10^{-5}$ \\ \hline
        CASCADE & $-1.97 \times 10^{-7}$ & $-1.69 \times 10^{-7}$ & $-4.24 \times 10^{-7}$ & $-1.08 \times 10^{-6}$ & $-7.01 \times 10^{-5}$ & $-8.01 \times 10^{-5}$ & $-9.42 \times 10^{-5}$ & $-6.69 \times 10^{-5}$ & $-2.82 \times 10^{-7}$ \\
        \noalign{\hrule height 1.5pt}
    \end{tabular}
    }
    \caption{Normalized log probabilities when there are $3$ modes.}
    \label{tab:3_modes}
\end{table}

\vfill

\begin{table}[ht]
    \centering
    \resizebox{0.99\linewidth}{!}{%
    \begin{tabular}{c!{\vrule width 1.25pt}c|c|c!{\vrule width 1pt}c|c|c|c|c|c}
        \noalign{\hrule height 1.5pt}
        & \fts\qts & \fwiki\qwiki & \fcode\qcode & \fts\qwiki & \fts\qcode & \fwiki\qts & \fwiki\qcode & \fcode\qts & \fcode\qwiki \\
        \noalign{\hrule height 1.25pt}
        Direct training ($r = \infty$) & $-6.73 \times 10^{-7}$ & $-9.03 \times 10^{-7}$ & $-6.25 \times 10^{-7}$ & $-4.19 \times 10^{-2}$ & $-0.277$ & $-5.54 \times 10^{-2}$ & $-1.39 \times 10^{-2}$ & $-0.172$ & $-3.80 \times 10^{-3}$ \\ \hline
        Full rewriting ($r = 1$) & $-5.68 \times 10^{-6}$ & $-5.80 \times 10^{-6}$ & $-6.25 \times 10^{-6}$ & $-2.21 \times 10^{-5}$ & $-2.89 \times 10^{-5}$ & $-6.16 \times 10^{-5}$ & $-5.51 \times 10^{-5}$ & $-3.12 \times 10^{-5}$ & $-1.65 \times 10^{-5}$ \\
        \noalign{\hrule height 1.5pt}
    \end{tabular}
    }
    \caption{Normalized log probabilities when using phi-1-small (350M).}
    \label{tab:3_modes_bigger_model}
\end{table}

\vfill

\end{landscape}

\end{document}